\pdfoutput=1

\documentclass[11pt]{article}

\usepackage[]{EMNLP2022}

\usepackage{times}
\usepackage{latexsym}
\usepackage{graphicx}       
\usepackage{booktabs}       
\usepackage{amsmath}        
\usepackage{multirow}

\usepackage[T1]{fontenc}

\usepackage[utf8]{inputenc}

\usepackage{microtype}

\usepackage{inconsolata}

\title{Enhancing Slot Tagging with Intent Features for Task Oriented Natural Language Understanding using BERT}

\author{Shruthi Hariharan \\ {\bf Vignesh Kumar Krishnamurthy} \\ {\bf Jayantha Gowda Sarapanahalli} \\
  Samsung R\&D Institute \\
  Bengaluru, India \\
  \texttt{\{s.hariharan,vig.kirsh,jayantha.g\}} \\ 
  \texttt{@samsung.com} \\\And
  Utkarsh \\
  Indian Institute of Technology \\
  Kanpur, India \\
  \texttt{rajpututkarsh530@gmail.com}}

\begin{document}
\maketitle
\begin{abstract}
Recent joint intent detection and slot tagging models have seen improved performance when compared to individual models. In many real-world datasets, the slot labels and values have a strong correlation with their intent labels. In such cases, the intent label information may act as a useful feature to the slot tagging model. In this paper, we examine the effect of leveraging intent label features through 3 techniques in the slot tagging task of joint intent and slot detection models. We evaluate our techniques on benchmark spoken language datasets SNIPS and ATIS, as well as over a large private Bixby dataset and observe an improved slot-tagging performance over state-of-the-art models.  
\end{abstract}

\section{Introduction}

With the growing popularity of Virtual Assistants, the importance of Spoken Language Understanding (SLU) has increased, and there are greater expectations placed by end users in their capabilities for fulfilling expectations. Such systems are expected to be robust in terms of understanding and carrying out user intent, with all necessary execution parameters identified correctly. These systems usually comprise multiple domains, and cater to executing one or more actions after a particular domain and intent are chosen based on natural language input from the user.

Given an input sentence, Spoken Language Understanding as implemented in a virtual assistant system requires the system to identify the set of action(s) to carry out (intent identification/classification), and the parameters required for execution (slot tagging), among other tasks \citep{survey}.

Intent identification is a sentence classification problem – where an input sentence passed through a classification model to predict a class after deriving features from it \cite{survey}. Slot tagging is a sequence-labelling problem – where a model assigns labels to contiguous spans of tokens in an utterance. This is typically represented in an IOB notation, where a B tag indicates the beginning of the span, I tag indicates the continuation of the span and O indicates the token is not of interest \cite{survey}. For example, the input sentence “What's the time in Mumbai?” could be assigned an intent label and its tokens can be tagged as slots as shown in Table \ref{tab:iob_format}, to indicate the function a virtual assistant is supposed to perform and the parameters it would require to carry it out.

\begin{table}[t]
  \centering
  \begin{tabular}{llllll}
    \toprule
    \multicolumn{1}{c}{\textbf{Query}}    & \multicolumn{1}{c}{What's}  & \multicolumn{1}{c}{the} & \multicolumn{1}{c}{time} & \multicolumn{1}{c}{in} & \multicolumn{1}{c}{Mumbai}\\
    
    \multicolumn{1}{c}{\textbf{Slots}}    & \multicolumn{1}{c}{O}  & \multicolumn{1}{c}{O}  & \multicolumn{1}{c}{O}  & \multicolumn{1}{c}{O}  & \multicolumn{1}{c}{B-City}  \\
    
    \multicolumn{1}{c}{\textbf{Intent}}   & \multicolumn{5}{c}{GetCurrentTimeAtLocation}  \\
    \bottomrule
  \end{tabular}
  \caption{Intent and slot labels assigned to an example utterance in IOB notation}
    \label{tab:iob_format}
\end{table}

Driven by the proliferation of ventures in several fields, such as personal assistants, chat bots, robotics etc., the joint task of identifying intents and slots has seen several approaches \cite{survey}. In this work, we examine 3 unique techniques of using intent features to bolster the information available for the slot-tagging task and demonstrate the improvement on it. Though the below methods are proposed to be used in a joint setting, they may also be used in an independent slot tagging task setting if the intent label can be guaranteed or arrived at using different means.

\section{Related Work}

Prior work performs correlations between intent and slot information through several techniques. One approach proposes an adaptive intent-slot graph structure to allow predicted multiple intents to combine with output information of a unidirectional LSTM to facilitate and improve slot predictions \cite{agif}.

Another approach builds a non-autoregressive parallelized joint model and a 2-pass mechanism, the first of which isolates the predicted B- slots and the second uses the identified B- slots to identify and correct tagging errors. In this process, the intent representation and slot tag representation are concatenated and processed to make it a joint model \cite{slotrefine}.

Another comparable approach builds an architecture on top of the JointBERT\cite{jointbert} intent prediction model, pools the [CLS] token output with the individual slot prediction output, and fuses the hidden state output of each individual token with the pooled intent features and individual word features looked up using a separate NLP module \cite{enriched_pretrained_transformers}.

These approaches typically rely on using the intent representation (i.e. BERT output representing the intent) after processing the output of a language model, such as the output of the [CLS] token, or applying mechanisms after slot-tagging unit outputs have been derived, to correlate the information. In this work, we propose an alternate way to include the intent label information as part of the slot-tagging process, which we detail in the section below.


\section{Proposed Approach}

The approach we present here relies on language models such as BERT. More specifically, we make use of the language model for the slot-tagging task, by using the predicted intent label itself as part of the input. We will briefly describe BERT and highlight the aspect that helps in our approach.

\subsection{BERT}

Bidirectional Encoder Representations from Transformers or BERT has proven to be a very robust language model that performs well in many tasks \cite{bert}. The key feature of BERT we attempt to leverage here is how the generated output can vary for the same input under different context.

The utterances are transformed into input sequences and passed to the pre-trained BERT language model, where the output of the [CLS] token is used to identify intents, and the sequence output is used to identify slots. Given a batch of sequences as input, the final hidden state (i.e. pooled) output of BERT would be of dimension:

\begin{equation} \label{eq:bert_pooled_output}
(batch\ size, max.\ seq.\ length, hidden\ size)
\end{equation}

Where $batch\ size$ refers to the number of training examples in the batch, $max.\ seq.\ length$ indicates the number of tokens per example that the model is trained over (including padding tokens) and $hidden\ size$ indicates the size of the output per token from the BERT model.

One method to use this process for slot filling task is to add a dense/linear layer on top of this to arrive at final slot predictions. The Cross Entropy Loss between the predicted and actual slots is calculated for optimization. This may also involve using a CRF layer \cite{jointbert}.

The slot prediction task per token of the input sequence is formulated as:

\begin{equation} \label{eq:slot_prediction_task}
 y^s = softmax(W_sh_n + b_s), n \in 1
\end{equation}

Here, $h_n$ refers to the BERT hidden state output for the $n^{th}$ token, and $W_s$ and $b_s$ refer to the parameters of the linear layer. By minimizing the cross entropy loss of this output, the model is trained to predict slots \cite{jointbert}.

\subsection{Augmenting Slot Filling}

\begin{figure*}
  \includegraphics[width=\linewidth]{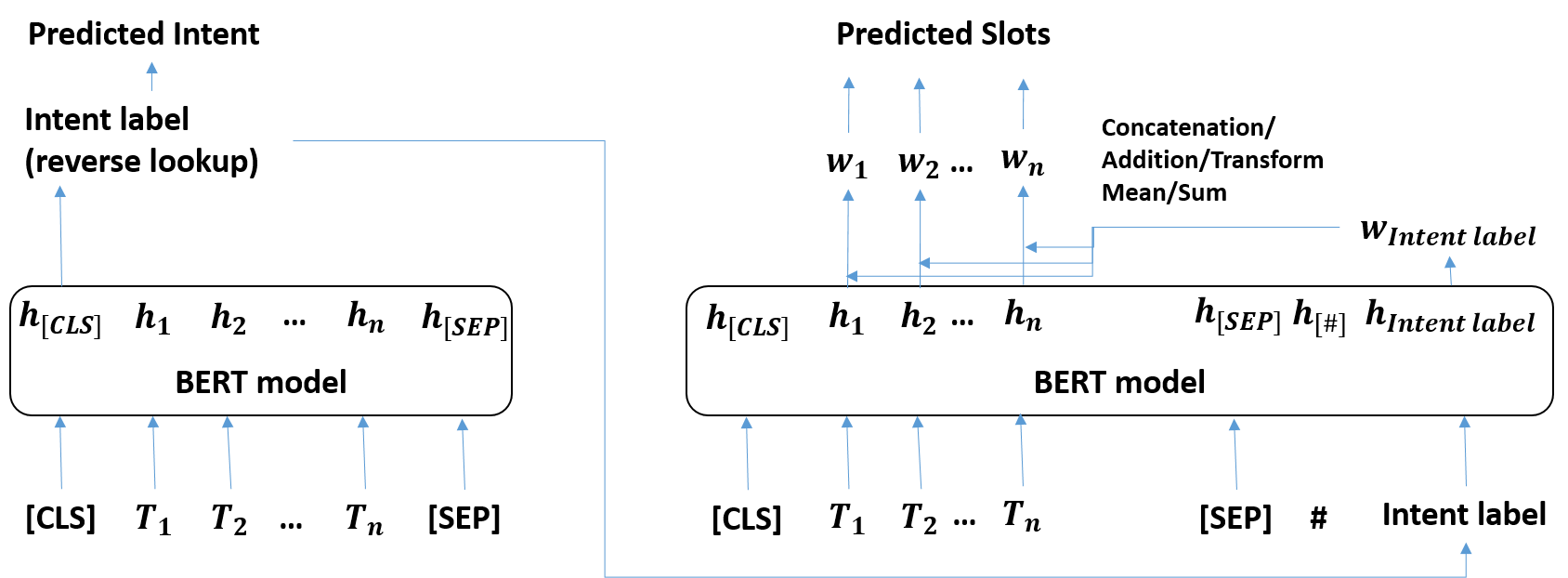}
  \caption{Initial steps of augmentation process.}
  \label{fig:architecture}
\end{figure*}

The proposed techniques described below were motivated by works on usage of intent information and correlation with slots \cite{agif} \cite{slotrefine} \cite{enriched_pretrained_transformers} a relation extraction model \cite{numerical_techniques_paper}, and a technique to incorporate labels as part of the task \cite{label_usage},

\begin{table}[b]
  \centering
  \begin{tabular}{ll}
    \toprule
    {\textbf{Label}}                & {SearchCreativeWork}\\
    {\textbf{BERT tokens}}      & {search \#\#cre \#\#ative \#\#work}\\
    {\textbf{CamelCase split}} & {search creative work}\\    
    \bottomrule
  \end{tabular}
  \caption{BERT tokenizer output formed from different intent label treatment}
  \label{tab:splitting_intent}
\end{table}

The initial set of steps described by the method is shown in Fig. \ref{fig:architecture}. After obtaining the intent label from a reverse lookup based on the [CLS] token output, we include the predicted intent label as part of the input sequence, by appending a concatenation of the ‘\#’ character and the intent label to the original utterance, to include it in the BERT-Tokenization process. This ensures that the embedding output generated is with the inclusion of the intent label. Tokenization of the intent label breaks up the intent into smaller comprehensible phrases for the model to associate with the slot prediction. As the BERT final hidden state output incorporates the intent information, we isolate it based on the position of the ‘\#’ character hidden state output and perform a dimensionality reduction on it as follows:

\begin{equation}
 \label{eq:conversion}
 \begin{aligned}
    (N_{chunks}, max.\ seq.\ length, hidden\ size)\\
    \to (1, max.\ seq.\ length, hidden\ size)
 \end{aligned}
\end{equation}

\begin{equation}
 \label{eq:intent_hidden_values}
 w^i_{intent} = [w^i_{intent_1},\ldots,w^i_{intent_{hidden\ size}}]
\end{equation}

\begin{equation}
 \label{eq:intent_values_operation}
 w_{intent} = \frac{\sum_iw^i_{intent}}{q}
\end{equation}

Given $w^i_{intent}$ to be the BERT model output of the intent label based on its tokenization, the reduction involves scaling the sum of $w$ for all chunks based on the value $q$. When $q = 1$, the result converges to the sum of all the values, and when $q=N_{chunks}$, the result converges to the mean value.

It is also noticeable that the intent labels can be comprised of multiple words concatenated, which can be individually identifiable. BERT, while processing will segment this label based on its WordPiece vocabulary \cite{wordpiece}. An alternative method is to split the words of the intent label on an identifiable basis, and perform the same process (Note that the BERT tokenizer may also segment the single words after the manual split).

By doing this, the embeddings produced based on the split can be considered as a more refined input compared to the unsplit and BERT-processed intent labels. An example of this is shown in the Table \ref{tab:splitting_intent}. In the first case, the intent label is used as the single word “SearchCreativeWork”, and in the second one, the intent label is split by camel case into three separate words “Search”, “Creative”, and “Work” (in both cases, before lowercasing and processing). 

Either intent representation can be used in the following three proposed methods to capture semantic relation between intents and slots:

\renewcommand{\arraystretch}{1.5}
\begin{tiny}
\begin{table*}[h]
\begin{tabular}{lcccc}
\hline \hline
\multirow{2}{*}{\textbf{Method}} & \multicolumn{2}{c}{\textbf{SNIPS}}    
& \multicolumn{2}{c}{\textbf{ATIS}}       \\
& \multicolumn{1}{c}{\textbf{Sem. Accuracy}} & \multicolumn{1}{c}{\textbf{F1 Score}} & \multicolumn{1}{c}{\textbf{Sem. Accuracy}} & \multicolumn{1}{c}{\textbf{F1 Score}}
\\
\hline \hline

JointBERT\cite{jointbert} & 92.8 & 97 & 88.2 & 96.1 \\
JointBERT with CRF\cite{jointbert}  & 92.6 & 96.7 & 88.6 & 96 \\
StackPropagation \cite{stack_prop}  & 92.9 & 97 & 88.6 & 96.1 \\
Slotrefine \cite{slotrefine} & 92.96 & 97.05 & 88.64 & 96.16 \\
Transformer-NLU \cite{enriched_pretrained_transformers} & 91.86 & 96.57 & 88.69 & 96.25 \\ \hline
Baseline & 92.86$^*$ & 97.07$^*$ & 87.46 & 96.02 \\
Concatenation w/ Sum & \textbf{94}$^+$$^*$ & \textbf{97.49}$^+$$^*$ & 86.79 & 95.26 \\
Concatenation w/ Mean & 92.43 & 97.02 & \textbf{88.69} & \textbf{96.26} \\ 
Addition w/ Sum & 93 & 97.13 & 87.46 & 95.85 \\
Addition w/ Mean & 92.86$^*$ & 97.18$^*$ & 88.47 & 96.01 \\
Transform w/ Sum & 92.71$^*$ & 97.13$^*$ & 87.68 & 95.87 \\
Transform w/ Mean & 93.57 & 97.07 & 88.35 & 96.06 \\                                
\hline \hline
 \end{tabular}
\caption{Results with SNIPS and ATIS Datasets. Intent accuracy is excluded as our work focuses solely on improving slot filling performance. For SNIPS results, + indicates CRF usage, and * indicates that the intent label was split by camel case.}
 \centering
 \label{tab:sota_results}
\end{table*}                                
\end{tiny}

\begin{enumerate}
 \item \textit{Concatenation}
 
 A larger dimension vector helps us capture intent and slot embedding together. The transformers capture the relation between slot tags and intention of the utterances. In this method, we transform the sequence embedding by appending the intent embedding to each token as shown below.
 
 \begin{equation} \label{eq:concat}
  \begin{aligned}
   (1, max\ seq.\ length, hidden\ size) \to\\
   (1, max\ seq.\ length, 2 * hidden\ size)
  \end{aligned}
 \end{equation}

 \item \textit{Addition}
 
 In this method, we add the reduced intent information to word embedding, which in essence transforms the result into a vector summation of both inputs
 
 \begin{equation} \label{eq:addition}
  w_f = w_{intent} + w_{word}
 \end{equation}
 
 \item \textit{Transformation}
 
 In this method, after performing dimensionality reduction, an appropriate function, which expresses the relation between $(w_{intent}$ and $w_{word})$ could be tuned by passing it to additional layers, such as a dense layer. Considering $w_f$ being the dimension-reduced output for example, the dense layer could perform the following operation:
 
 \begin{equation}
  y = softmax(xw_f + b)
 \end{equation}

 This adds a separate set of tunable parameters, which helps us capture relations between intent and utterance.

\end{enumerate}

\section{Experiments And Analysis}
The model has been primarily trained and tested on both the uncased and cased SNIPS dataset \cite{snips_dataset}, a voice assistant based dataset. The dataset has 14484 utterances, split into 13,084 training, 700 validation and 700 testing utterances. It contains 72 slots and 7 intents. Our primary focus was on this dataset as it is a better representative of a task oriented SLU system's capablities.

The model was also trained and tested on the ATIS dataset \cite{atis_dataset}, containing 4478 training, 500 validation and 893 test utterances. It contains 119 slots and 21 intents.

Additionally, the model was trained and tested on a private Bixby dataset of ~9000 utterances in the Gallery domain, containing 16 intents and 46 slots representing various Gallery application related functionalities.

\subsection{Training Details}
The modified models were primarily trained and tested with the “bert-large-uncased” model for the mentioned datasets and ``bert-large-cased'' model for the cased SNIPS dataset, with a hidden size of 1024, 16 attention heads and 24 hidden layers. The maximum number of words per utterance was chosen to be 50 for SNIPS and 70 for ATIS based on the respective dataset measures.

The SNIPS uncased model was trained for 10 and 15 epochs with learning rates 4e-5 and 5e-5, and dropout 0.1 and 0.2, including alternating between the intent label split based on camel case and use of a CRF layer. The training batch size was set as 32, and eval batch size 64. The model takes 1.5-2 hours to train on an NVIDIA P40 GPU.
For the ATIS dataset, the model was trained for 15 epochs, with a learning rate of 4e-5, dropout of 0.1 and usage of CRF layer. The best model was chosen for each method based on the dev set F1 score. The sentence level semantic frame accuracy is also considered for correctness, where the correct intent label must be predicted and all input tokens must be assigned the correct slot labels without missing or incorrect predictions \cite{semantic_frame_reference}. 

\subsection{Results}
We show the results indicating the semantic frame accuracy and the slot-F1 score in Tables  \ref{tab:sota_results} and \ref{tab:bixby_results}. The intent accuracy is not mentioned here as the focus of the work is on improving slot tagging. Bolded numbers indicate highest values. Here, the values shown as ``Baseline'' indicate the values obtained from the JointBERT model \citep{jointbert} by adding the intent label to the utterances without augmentation. The subsequent rows show the results upon including the proposed techniques. The SNIPS dataset has entities like album and song names that are better distinguished in their cased forms, and the results for the cased model are shown in the appendix in table \ref{tab:cased_snips_results}.

We observe that all of these techniques demonstrate an increased or comparable performance in the semantic frame accuracy as well as in the slot F1 score. The same effect was seen with other private datasets as well - in fact, the improvement in performance was much more substantial in the case of private datasets.

Though we show results for ATIS dataset in this work, the slots are mostly independent of the intents. The same slots are reused across utterances for multiple intents; for instance, we found atleast 8 intents using the same slot label \textit{toloc.city\_name} and \textit{fromloc.city\_name} (Refer to Table \ref{tab:atis_eg}). For this reason, the effect of the augmentation techniques is less pronounced \cite{is_atis_shallow}.

\subsection{Dataset Error Analysis}
While analysing the failures of our model, we observed several annotation and intent errors in the SNIPS train and test sets. This led us to go through the entire test set and make corrections wherever there were clear errors in the test cases. These corrections have been detailed in the Appendix Tables \ref{tab:snips_test_spelling_mistakes} - \ref{tab:snips_test_wrong_slots_3}. We re-ran our models on the corrected test set, and also ran the models for \citep{jointbert}, \citep{slotrefine} and \citep{stack_prop} for which source code was available. The results have been tabulated in Appendix Table \ref{tab:updated_dataset_uncased_results}. Similar corrections were made on the cased dataset as well.

An observation we can draw from these tabulated results is that the cased BERT model recognizes named entities a little better due to the casing of the words in the utterance, and thus shows improved performance for SNIPS dataset, as compared to the uncased model. This may not necessarily be due to the model recognizing the cased entity for what it exactly is, but rather due to other factors such as patterns or the intent label. Most of the other errors involved confusions between similar named entities like album, artist, and song names. This may be because the names are unseen by the BERT vocabulary and training data, and similar sentence patterns exist in training involving all three entities. A scalable solution to this may be to use an external data source to distinguish the correct entity type.

We also observed errors in the training dataset such as the phrase ``italian-american cuisine'' being tagged as ``served\_dish'' rather than the expected ``cuisine'', and also inconsistencies such as ``play \_\_\_ song'' being assigned the ``SearchCreativeWork'' intent label rather than ``PlayMusic''. We also observed an inconsistency in tagging the word 'the' as part of the named entity, that confuses the model when encountering the word in uncased text. Due to the large size of the train dataset, the correction of the train set is out of the work's scope, and to maintain consistency across other research papers, we limit the corrections to only the test dataset.

\begin{footnotesize}
\renewcommand{\arraystretch}{1.5}
\begin{table}[t]
\centering
\begin{tabular}{lcc}
\hline
\textbf{Method} & \textbf{SFA} & \textbf{F1 Score} \\ \hline
Baseline (w/ Intent Label) & 90.9 & 97.8 \\
Concatenation w/ Mean & \textbf{93.3} & \textbf{98.3} \\
Concatenation w/ Sum & 92.2 & 97.8 \\
Addition w/ Sum & 90.8 & 97.8 \\
Addition w/ Mean & 92.6 & 97.7 \\
Transformation w/ Sum & 90.5 & 97 \\
Transformation w/ Mean & 92.1 & 97.7 \\
\hline
\end{tabular}
\caption{Results on Bixby Gallery dataset with proposed techniques and baseline}
\label{tab:bixby_results}
\end{table}
\end{footnotesize}
\renewcommand{\arraystretch}{1}

\section{Future Work}
There may be scope for improvement based on modifications like using a different language model in place of BERT, or using different word embeddings, as the process itself does not change upon changing the language model. Other experiments may include adding a more sophisticated layer in the Transformation technique mentioned in section 3, fine-tuning the language model on the domain-specific vocabulary, or using other means to resolve entities in language model. There is also scope for extending this to a multi-label multi-class classification problem. The presented techniques do not differ based on the language of the text, so they can be experimented upon with other text languages.

\bibliography{references}
\bibliographystyle{acl_natbib}
\onecolumn
\appendix

\section{Appendix}
\label{sec:appendix}


\begin{table*}[h]
  \centering
  \begin{tabular}{llllllllll}
    \hline \hline    
	
    \multicolumn{1}{c}{\textbf{Intent}}   & \multicolumn{7}{c}{\textbf{atis\_flight}}  \\
    \multicolumn{1}{c}{\textbf{Query}}    &  \multicolumn{1}{c}{give} & \multicolumn{1}{c}{me} & \multicolumn{1}{c}{flights} & \multicolumn{1}{c}{from} & \multicolumn{1}{c}{pittsburgh} & \multicolumn{1}{c}{to} & \multicolumn{1}{c}{baltimore} \\

    \multicolumn{1}{c}{\textbf{Slots}}   & \multicolumn{1}{c}{O} & \multicolumn{1}{c}{O} & \multicolumn{1}{c}{O} & \multicolumn{1}{c}{O} & \multicolumn{1}{c}{B-fromloc.city\_name} & \multicolumn{1}{c}{O} & \multicolumn{1}{c}{B-toloc.city\_name}  \\

	\hline

	\multicolumn{1}{c}{\textbf{Intent}}   & \multicolumn{7}{c}{ \textbf{atis\_airline}}  \\
    \multicolumn{1}{c}{\textbf{Query}}    & \multicolumn{1}{c}{what}   & \multicolumn{1}{c}{airlines}   & \multicolumn{1}{c}{fly}   & \multicolumn{1}{c}{between}   & \multicolumn{1}{c}{boston} & \multicolumn{1}{c}{and}   & \multicolumn{1}{c}{atlanta} \\
    \multicolumn{1}{c}{\textbf{Slots}}   & \multicolumn{1}{c}{O}   & \multicolumn{1}{c}{O}   & \multicolumn{1}{c}{O}   & \multicolumn{1}{c}{O}   & \multicolumn{1}{c}{B-fromloc.city\_name}   & \multicolumn{1}{c}{O}   & \multicolumn{1}{c}{B-toloc.city\_name} \\

	\hline \hline
  \end{tabular}
  \caption{An example of similar annotations for 2 different intents in ATIS.}
  \label{tab:atis_eg}
\end{table*}

\begin{footnotesize}
\renewcommand{\arraystretch}{1.5}
\begin{table*}[h]
 \centering
\begin{tabular}{lcc}
 \hline
 \textbf{Method} & \textbf{Semantic Frame Accuracy} & \textbf{F1 Score} \\ \hline
 Baseline (w/intent label) & 94.13 & 97.57 \\
 Concatenation w/ Sum & 93.71 & 97.21 \\
 Concatenation w/ Mean & 94.13 & 97.13 \\
 Addition w/ Mean & 94.42 & 97.43 \\
 Addition w/ Sum & 94.28 & 97.32 \\
 Transformation w/ Sum & \textbf{96.57} & 97.71 \\
 Transformation w/ Mean & 95.42 & \textbf{97.84} \\
 \hline
\end{tabular}
\caption{Results on corrected cased SNIPS dataset with proposed techniques and baseline}
\label{tab:cased_snips_results}
\end{table*}
\end{footnotesize}
\renewcommand{\arraystretch}{1.3}

\begin{footnotesize}
\renewcommand{\arraystretch}{1.5}

\begin{table*}[h]
\centering
\begin{tabular}{lcc}
\hline
\textbf{Method} & \textbf{Semantic Frame Accuracy} & \textbf{F1 Score} \\ \hline
JointBERT (with CRF) \citep{jointbert} & 92.71 & 96.96 \\
SlotRefine (w/o Glove/BERT) \citep{slotrefine} & 94.49 & 85.7 \\
StackPropagation (w/o BERT) \citep{stack_prop} & 85.85 & 93.76 \\
\hline
Baseline (w/ Intent label) & 93.71 & 97.19 \\
Concatenation w/ Mean & 93.86 & 97.46 \\
Concatenation w/ Sum & \textbf{95.29} & \textbf{97.94} \\
Addition w/ Sum & 94 & 97.54 \\
Addition w/ Mean & 93.71 & 97.38 \\
Transformation w/ Sum & 93.29 & 97.18 \\
Transformation w/ Mean & 94.14 & 97.35 \\
\hline
\end{tabular}
\caption{Results on the updated uncased SNIPS dataset with proposed techniques and baseline}
\label{tab:updated_dataset_uncased_results}
\end{table*}
\end{footnotesize}
\renewcommand{\arraystretch}{1.3}

\begin{table*}[h]
 \centering
 \begin{tabular}{l}
  \hline
  add this album \textbf{ny} bill callahan to my mi casa es la tuya playlist oficial list \\
  add this album \textbf{by} bill callahan to my mi casa es la tuya playlist oficial list \\
  \hline
  play the last wellman braud album \textbf{relaesd} \\
  play the last wellman braud album \textbf{released} \\
  \hline
  i need the \textbf{wather} for next week in the philippines \\
  i need the \textbf{weather} for next week in the philippines \\
  \hline
 \end{tabular}
 \caption{SNIPS uncased test set utterances with spelling mistakes}
 \label{tab:snips_test_spelling_mistakes}
\end{table*}

\begin{table*}[h]
 \centering
 \begin{tabular}{lll}
  \hline
  \textbf{Utterance} & \textbf{Original Intent} & \textbf{Corrected Intent} \\
  \hline
  find on dress parade & SearchScreeningEvent & SearchCreativeWork \\
  play the new noise theology e p & SearchCreativeWork & PlayMusic \\
  find now and forever & SearchScreeningEvent & SearchCreativeWork \\
  i need a table in uruguay in 213 days when it s chillier & GetWeather & BookRestaurant \\
  find heat wave & SearchScreeningEvent & SearchCreativeWork \\
  play the electrochemical and solid state letters song & SearchCreativeWork & PlayMusic \\
  play the song memories are my only witness & SearchCreativeWork & PlayMusic \\
  i want to listen to the song only the greatest & SearchCreativeWork & PlayMusic \\
  find the panic in needle park & SearchScreeningEvent & SearchCreativeWork \\
  play the album journeyman & SearchCreativeWork & PlayMusic \\
  what is dear old girl cooper foundation & SearchScreeningEvent & SearchCreativeWork \\
  \hline
 \end{tabular}
 \caption{SNIPS uncased test set utterances with incorrectly labeled intents}
 \label{tab:snips_test_wrong_intents}
\end{table*}

\begin{table*}[h]
 \centering
 \begin{tabular}{p{\linewidth}}
 \hline
  put lindsey cardinale into my hillary clinton s women s history month playlist   \\ O B-artist I-artist O B-playlist\_owner B-playlist I-playlist I-playlist I-playlist I-playlist I-playlist I-playlist I-playlist \\ O B-artist I-artist O B-playlist\_owner B-playlist I-playlist I-playlist I-playlist I-playlist I-playlist I-playlist O \\ 
\hline
find on dress parade \\ O B-movie\_name I-movie\_name I-movie\_name \\ O B-object\_name I-object\_name I-object\_name \\ 
\hline
add the song don t drink the water to my playlist \\ O O B-music\_item B-playlist I-playlist I-playlist I-playlist I-playlist O B-playlist\_owner O \\ O O B-music\_item B-entity\_name I-entity\_name I-entity\_name I-entity\_name I-entity\_name O B-playlist\_owner O \\ 
\hline
add to the rock games \\ O O O B-playlist B-entity\_name \\ O O O B-playlist I-playlist \\ 
\hline
play the new noise theology e p  \\ O B-object\_name I-object\_name I-object\_name I-object\_name I-object\_name I-object\_name \\ O O B-album I-album I-album I-album I-album \\ 
\hline
add lisa m to my guitar hero live playlist \\ O B-artist I-artist O B-playlist\_owner B-playlist I-playlist I-playlist I-playlist \\ O B-artist I-artist O B-playlist\_owner B-playlist I-playlist I-playlist O \\ 
\hline
find now and forever \\ O B-movie\_name I-movie\_name I-movie\_name \\ O B-object\_name I-object\_name I-object\_name \\ 
\hline
book a bar that serves italian-american cuisine neighboring wilson av for one person  \\ O O B-restaurant\_type O O B-served\_dish I-served\_dish B-spatial\_relation B-poi I-poi O B-party\_size\_number O \\ O O B-restaurant\_type O O B-cuisine O B-spatial\_relation B-poi I-poi O B-party\_size\_number O \\ 
\hline
book a restaurant for 3 people at eighteen oclock in saint vincent and the grenadines  \\ O O B-restaurant\_type O B-party\_size\_number O O B-timeRange O O B-country I-country I-country I-country I-country \\ O O B-restaurant\_type O B-party\_size\_number O O B-timeRange I-timeRange O B-country I-country I-country I-country I-country \\ 
\hline
what is dear old girl cooper foundation \\
O O B-movie\_name I-movie\_name I-movie\_name B-location\_name I-location\_name \\
O O B-object\_name I-object\_name I-object\_name I-object\_name I-object\_name \\
\hline
\end{tabular}
 \caption{SNIPS test set utterances with wrong and corrected slot labels (Pt. 1). The format is utterance, followed by existing tagging and corrected tagging.}
 \label{tab:snips_test_wrong_slots_1}
 \end{table*}
\begin{table*}[h]
 \centering
 \begin{tabular}{p{\linewidth}}
 \hline
 play the last wellman braud album released  \\ O O O B-artist I-artist B-music\_item O \\ O O B-sort B-artist I-artist B-music\_item O \\ 
\hline
i need a table in uruguay in 213 days when it s chillier \\ O O O O O B-country B-timeRange I-timeRange I-timeRange O O O B-condition\_temperature \\ O O O O O B-country B-timeRange I-timeRange I-timeRange O O O O \\ 
\hline
find heat wave \\ O B-movie\_name I-movie\_name \\ O B-object\_name I-object\_name \\ 
\hline
play the electrochemical and solid state letters song \\ O O B-object\_name I-object\_name I-object\_name I-object\_name I-object\_name B-object\_type \\ O O B-track I-track I-track I-track I-track B-music\_item \\ 
\hline
play the song memories are my only witness \\ O O B-object\_type B-object\_name I-object\_name I-object\_name I-object\_name I-object\_name \\ O O B-music\_item B-track I-track I-track I-track I-track \\ 
\hline
i want to listen to the song only the greatest \\ O O O O O O B-object\_type B-object\_name I-object\_name I-object\_name \\ O O O O O O B-music\_item B-track I-track I-track \\ 
\hline
is it going to be chilly in western sahara in 13 hours \\ O O O O O B-condition\_temperature O B-country I-country O B-timeRange I-timeRange \\ O O O O O B-condition\_temperature O B-country I-country B-timeRange I-timeRange I-timeRange \\ 
\hline
i want to go to 88th st-boyd av or close by and book seats for 10 \\ O O O O O B-poi I-poi I-poi O B-spatial\_relation O O O O O B-party\_size\_number \\ O O O O O B-poi I-poi I-poi O B-spatial\_relation I-spatial\_relation O O O O B-party\_size\_number \\ 
\hline
show me movie time for i am sorry  at my movie house \\ O O O O O B-movie\_name I-movie\_name I-movie\_name O O B-object\_location\_type I-object\_location\_type \\ O O B-object\_type I-object\_type O B-movie\_name I-movie\_name I-movie\_name O O B-object\_location\_type I-object\_location\_type \\ 
\hline
 \end{tabular}
 \caption{SNIPS test set utterances with wrong and corrected slot labels (Pt. 2)}
 \label{tab:snips_test_wrong_slots_2}
\end{table*}

\begin{table*}[h]
 \centering
 \begin{tabular}{p{\linewidth}}
 
\hline
i would like to book a highly rated brasserie with souvlaki neighboring la next week  \\ O O O O O O B-sort I-sort B-restaurant\_type O B-cuisine B-spatial\_relation B-state B-timeRange I-timeRange \\ O O O O O O B-sort I-sort B-restaurant\_type O B-served\_dish B-spatial\_relation B-state B-timeRange I-timeRange \\ 
\hline
find the panic in needle park \\ O B-movie\_name I-movie\_name I-movie\_name I-movie\_name I-movie\_name \\ O B-object\_name I-object\_name I-object\_name I-object\_name I-object\_name \\ 
\hline
play the album journeyman \\ O O B-object\_type B-object\_name \\ O O B-music\_item B-album \\ 
\hline
i want to add a song by jazz brasileiro \\ O O O O O B-music\_item O B-playlist I-playlist \\ O O O O O B-music\_item O B-artist I-artist \\ 
 \hline
 book a table for chasity ruiz and mary at the fat duck in puerto rico \\
O O O O B-party\_size\_description I-party\_size\_description I-party\_size\_description I-party\_size\_description O B-restaurant\_name I-restaurant\_name I-restaurant\_name O B-country I-country \\
O O O O B-party\_size\_description I-party\_size\_description I-party\_size\_description I-party\_size\_description O B-restaurant\_name I-restaurant\_name I-restaurant\_name O B-state I-state \\
\hline
\end{tabular}
 \caption{SNIPS test set utterances with wrong and corrected slot labels (Pt. 3)}
 \label{tab:snips_test_wrong_slots_3}
\end{table*}
\end{document}